# Discrete Modeling of Multi-Transmitter Neural Networks with Neuron Competition

Nikolay Bazenkov[1], Varvara Dyakonova[2], Oleg Kuznetsov[1],
Dmitri Sakharov[2], Dmitry Vorontsov[2],
Liudmila Zhilyakova[1]

[1]*Trapeznkov Institute of Control Sciences of RAS, Moscow, Russian Federation*
[2] *Koltzov Institute of Developmental Biology of RAS, Moscow, Russian Federation*
n.bazenkov@yandex.ru, D.Vorontsov@idbras.ru, V.Dyakonova@idbras.ru, zhilyakova@ipu.ru,
olpkuz@yandex.ru, dant1930@gmail.com

**Abstract**
We propose a novel discrete model of central pattern generators (CPG), neuronal ensembles generating rhythmic activity. The model emphasizes the role of nonsynaptic interactions and the diversity of electrical properties in nervous systems. Neurons in the model release different neurotransmitters into the shared extracellular space (ECS) so each neuron with the appropriate set of receptors can receive signals from other neurons. We consider neurons, differing in their electrical activity, represented as finite-state machines functioning in discrete time steps. Discrete modeling is aimed to provide a computationally tractable and compact explanation of rhythmic pattern generation in nervous systems. The important feature of the model is the introduced mechanism of neuronal competition which is shown to be responsible for the generation of proper rhythms. The model is illustrated with two examples: a half-center oscillator considered to be a basic mechanism of emerging rhythmic activity and the well-studied feeding network of a pond snail. Future research will focus on the neuromodulatory effects ubiquitous in CPG networks and the whole nervous systems.

## 1 Introduction

Human knowledge about the diversity of neurotransmitters and their receptors provides a solid foundation for neuro- and psychopharmacology. Nevertheless, the chemical heterogeneity inherent to natural neuronal populations remains beyond the formalization that still treats the brain as an instance of electrical wiring diagram. We intend to overcome this, in some cases crucial, reduction.

One of the main objectives of the study is to show the informational significance of the chemical composition of the extracellular space (ECS) and to emphasize the important role of non-synaptic chemical interactions in the formation of coordinated electrical activity of neurons and neuromodulating effects. To identify this role explicitly, we investigate the non-synaptic mechanism of neural communication implemented by the multitransmitter interactions (so-called volume transmission). This is a conscious idealization, which is designed to simplify the modeling of broadcast signals in nervous systems.

Another objective is to justify the choice of a discrete approach for modeling the multitransmitter interactions, and construction of a viable model, which would possess an explanatory and predictive power.

Combining both objectives, we propose a *heterogeneous active neural network*. It consists of a set of active elements interacting according to specified rules, which dynamically determine the topology of the network, depending on the chemical composition of the ECS. It will be shown that changes in the content of transmitters can quickly modify or completely rebuild the topology of active networks. The proposed mathematical apparatus corresponds to biological realities; it explains and predicts the rapid processes of reconfiguration of neural networks resulting in different output rhythmic activities. This phenomenon can hardly be adequately described in models with rigid connections, passive elements, and binary input and output signals.

## 2 Survey of mathematical modeling of neurons and neural networks

At present, a wide variety of mathematical models describing individual neurons and networks of interacting neurons exists. The whole diversity of models can be divided into two large classes, which use the opposite approaches.

The first class includes the so-called continuous models describing biological neurons and their membrane processes in the form of systems of differential equations. The second class – discrete models – mostly consists of artificial neural networks of different sorts.

### 2.1 Continuous models

One of the earliest models, "integrate-and-fire", was proposed by L. Lapicque (Abbott, 1999) in 1907. It contains one differential equation describing the increase of membrane potential to a threshold value, then the emergence of a spike (or action potential), and then the regression of the membrane potential to the residual value. The Hodgkin–Huxley model (Hodgkin & Huxley, 1952), the best known popular model, which describes the occurrence and propagation of the action potential along the axon, was developed in 1952. The authors were awarded the Nobel Prize in Physiology and Medicine (1963). This model is a system of four differential equations containing a large number of parameters. One of its simplifications is the FitzHugh–Nagumo model (FitzHugh, 1969) (Nagumo, Arimoto, & Yoshizawa, 1962). The Morris–Lekar model (1981) is a combination of Hodgkin-Huxley and FitzHugh–Nagumo models describing the dynamics of calcium and potassium channels (Morris & Lecar, 1981). Besides, several different complications of "integrate-and-fire" model are worth noting: the Hindmarsh–Rose model, which can be considered as elaboration of the FitzHugh–Nagumo model, and a set of several multi-compartment models, describing different parts of a neuron (soma, axon, dendrites, and their interactions). In (Vavoulis, et al., 2007) the two-compartment model of neuron based on Hodgkin–Huxley model is created as an element of a computational model of the central pattern generator circuit.

A separate area of research contains the models that take into account the periodic endogenous activity of neurons – oscillatory neural networks. Such networks function due to the coordinated interaction of oscillating elements (Tsukerman & Kulakov, 2015) (Hyafil, Fontolan, Kabdebon, Gutkin, & Giraud, 2015).

In addition to the models describing the electrical activity of neurons there are models of transmitter interactions between neurons. In the model of Koch and Segev (Koch & Segev, 1999) two transmitters (glutamate and GABA) and four types of receptors (excitatory and inhibitory to each transmitter) are proposed. The transmitters affect the activity of neurons in different ways. The response of neurons to individual neurotransmitters is modeled by the extension of the Hodgkin–Huxley model.

A review of continuous computational models of neurons, neuron compartments, and neural networks can be found in the book (Sterratt, Graham, Gillies, & Willshaw, 2011).

The main advantage of these models is their expressive power – they describe the processes taking place on a cellular membrane with a high degree of accuracy. However, this advantage turns into a disadvantage: an abundance of parameters, some of which cannot be measured accurately, makes the model unstable to the initial data and parameters values, which often need to be tuned manually. As a consequence, there are two more drawbacks: a huge computational complexity, rapidly growing with the increase in the number of interacting neurons, and the lack of scalability. Therefore, such models can describe only networks with a small number of neurons. E.g. in the paper (Vavoulis, et al., 2007) a computational model of the feeding central pattern generator (CPG) of a snail is constructed; the model of the bursting neuron developed in (Ghigliazza & Holmes, 2004) was used in (Ghigliazza & Holmes, 2004) to construct the interaction of interneurons in CPG and motor neurons in the locomotor system of insects; in the model (Roberts, Conte, Buhl, Borisyuk, & Soffe, 2014), equations of the Hodgkin–Huxley type are used to describe the neurons of the tadpole, germinating axons and forming new synaptic connections. Networks in all these models contain rather small number of interacting neurons.

Continuous models claim to describe the nervous system as a whole, using methods of nonlinear dynamics. The paper (Rabinovich, Varona, Selverston, & Abarbanel, 2006) contains an exposition of the basic concepts and principles of this direction, as well as an overview of some modeling results.

## 2.2 Discrete models

In contrast to continuous models, discrete models tend to formalize intrinsic and extrinsic neuronal processes to make them as simple as possible. The simplification is achieved due to the parameter discretization, the scale-up of events and the ignoring of many informationally insignificant details. The most common types of discrete models are the artificial (formal) neurons and neural networks.

For the first time, a formal neuron and artificial neural network (ANN) were proposed by McCulloch and Pitts in 1943 (McCulloch & Pitts, 1943). In (Kleene, 1956) Kleene showed that networks of McCulloch–Pitts elements are equivalent to finite automata. This article initiated active research of ANNs.

The first application of ANNs for the intellectual process modeling was the perceptron, proposed by Rosenblatt in 1957 and described in the book (Rosenblatt, 1962). Research of Rosenblatt caused a significant growth of further investigations, but also greatly inflated expectations. However, after M. Minsky and S. Papert (Minsky & Papert, 1969) showed the principle limitations of perceptrons, the interest in ANNs significantly decreased for a long time. It was reborn only in the late 70's, and is preserved to the present day. For a detailed review of the models of formal neurons and ANNs see the book (Haykin, 2009).

The ongoing improvement of ANNs is stimulated mostly due to the success of intelligent technologies and the growth of their applications. In particular, the algorithms of deep learning of multilayer neural networks are developing rapidly (LeCun, Bengio, & Hinton, 2015), (Goodfellow, Bengio, & Courville, 2016), where "depth" means the number of layers of the network. The range of problems being solved by such networks is extremely wide due to their high ability to solve many types of poorly structured tasks: image, speech and handwriting recognition, classification, diagnostic tasks, computer vision, natural language processing, and even creative tasks.

Since ANNs are distributed computing architectures, they do not pretend to be models of biological neurons and nervous systems in general. Therefore, until the 2000s, neuroscientists didn't show any significant interest to ANNs and discrete models in general. The situation changed dramatically with the advent of a new discrete formalism "complex networks" (Dorogovtsev, 2010) (Kuznetsov, 2015), which was proved to be an adequate tool to describe the structures of various biological, technical and social systems. This formalism gave the first opportunity to consider the real neural system at the macro level, i.e. to investigate their global characteristics: the degree of clustering, the average length of paths etc. So, the appearance of complex networks attracted the attention of neuroscientists and led to the emergence of the concept of "structural and functional connectomics" (Baronchelli, Ferrer-i-Cancho, Pastor-Satorras, Chater, & Christiansen, 2013) (Bullmore & Sporns, 2009), which considers the nervous system as a large complex network – "connectome". Total structural connectome with nodes-neurons (all anatomical connections of the whole nervous system) to date, built only for the worm C. elegans, an organism which possesses only 302 neurons (Jarrell, et al., 2012). However, the hope that a complete connectome of the worm would be sufficient for understanding its behavior did not materialize (Bargmann, 2012).

One more type of models is represented in (Roberts P. , 1998) where the automata-based language is used for the description of real neural systems. In that study, as in artificial neurons of McCulloch – Pitts, all the complexity of cellular mechanisms is aggregated in binary states. Important feature of this model is the use of neurons with several types of endogenous activity: bursting neurons, neurons with tonic activity, neurons-triggers.

## 2.3 Types of models and our approach

Continuous and discrete models describe various aspects of the functioning of real neurons and thereby complement each other. In terms of continuous models, physico-chemical properties and processes in neurons and neural networks, like gradual changes in the membrane potential, action potentials, synaptic events etc., can be described quite accurately. However, many of the parameters of these descriptions are not essential for describing the information processes in the nervous systems – just as for understanding the operation of a computer it is not necessary to know the physical processes occurring in its elements and schemes. From our point of view, the parameters essential for the description of informatics of the nervous system, are the following:

- Inputs and outputs of neurons, of neural ensembles, of networks formed by ensembles, etc. (in terms of transmitters and receptors);
- The structure of connections (electrical and chemical) between neurons and ensembles;
- Possible input and output signals (in terms of electrical signals and neurotransmitters);
- Nature of the dependence of the output signals on the input ones (taking into account different types of neurons);
- Time sequences of signals and description of patterns in these terms;
- Memory in neurons and ensembles (in terms of discrete states, varying thresholds and weights of connections).

The first two kinds of parameters (inputs, outputs and connections) characterize the structural (static) properties of nervous systems. The graph theory is an adequate language for their modeling. The remaining parameters characterize the functional (dynamic) properties. Novel approaches will be required for their description.

To realize our goals, the existing discrete models need substantial expansions in several directions.

1. In all above-described discrete models, neurons are connected by rigid "wire connections". Therefore, the network topology cannot undergo rapid changes and rearrangements. Such phenomena as segregation of active subnetworks (ensembles) from larger networks, changing of rhythmic activity, rearrangement of central pattern generators observed in living systems need a fundamentally different language of description. The approach we develop extends the notion of a neural network: the connections in the proposed model denote not only pairwise synaptic connections, but also transmitter interactions of many neurons via the common extracellular space (ECS). These interactions can be much more variable. Thus, the structure of the connections can be dynamically rearranged and thereby the neuromodulation can be realized.

2. Artificial neural networks have very little in common with biological neural ensembles, since the former are constructed from the same elements, and in the latter the elements are functionally different. This difference, in our opinion, lies in the very basis of the nervous system, and is not an optional property of neurons. Models of neural networks must be heterogeneous in at least two respects. First, they must reflect the informationally significant diversity of transmitters and receptors sensitive to them. Interactions between neurons are not binary, and the model from "black&white" turns into "color"; the traditional language of logical (Boolean) networks (Kleene, 1956), (Wang & Albert, 2013), (Burks & Wright, 1953) is no longer suitable. Second, the model should include neurons with different types of activity.

# 3 Multi-neurotransmitter neuronal ensembles: biological background

The neurotransmitter (NT) diversity is a feature common to natural neuron populations. Further, the same canonical NTs, such as acetylcholine, dopamine, gamma-aminobutyric acid, histamine, octopamine and serotonin, are present in all or nearly all nervous systems, including the most primitive ones. Though not fully understood, this similarity indicates a fundamental role played by theneuron heterogeneity. Beginning from 1960s, speculations concerning the functional significance and evolutionary origins of multiple neuron phenotypes have been discussed in a number of theoretical and review papers (Florey, 1967), (Sakharov, 1974), (Bloom, 1984), (Getting, 1989), (Sakharov, 1990), (Dyakonova, 2012), (Moroz & Kohn, 2016).

While being most important for neuro- and psychopharmacology, multi-NT mechanisms have long been neglected by the theory of natural and artificial neural networks. As an approach to modeling multi-NT operations, the organization of biological neuronal ensembles is briefly outlined below. A neuronal ensemble may be defined as a group of nerve cells collectively involved in a particular neural computation. The empirical generalizations listed below are mostly derived from knowledge of small neural networks that produce appropriately ordered motor outputs even in the absence of input (the so called Central Pattern Generators, CPGs) (Meyrand, Simmers, & Moulins, 1991), (Mulloney & Smarandache, 2010), (Brezina, 2010), (Harris-Warrick, 2011), (Balaban, et al., 2015), (Marder, Goeritz, & Otopalik, 2015). It is supposed that, for cognitive functions, there may be "cognitive pattern generators", analogs of the familiar generators of motor commands (Graybiel, 1997).

1. Neurons can be recruited into an ensemble either permanently or transienty, from time to time, the latter mode depending on the composition of the ECS.

2. Neurons that constitute an ensemble utilize NTs for contact and distant communication. The same membrane receptor proteins are used in contact and distant NT effects, which are thus just different ends of a single spectrum rather than two distinct mechanisms. Contacts (the synapses) are mostly intended for rapid (phasic) actions of individual NT substances. Conversely, relatively slow (tonic, sustained) effects of individual or mixed NTs are exerted extrasynaptically. Neurotransmission without a synapse is called "volume transmission". The volume transmission is workable within the diffusion radius of the NT molecule.

3. Just as every hormone integrates organs of the body in its own way, a particular NT of a neuronal ensemble may act synergistically on its multiple cellular and molecular targets, thus integrating their individual responses into a specific and well-coordinated whole.

4. The effect of a neurotransmitter mixture is not equal to the sum of their individual effects. It seems that behind this generalization lies a reserve for increasing the information capacity of neural ensemble.

5. Generally, each neuronal ensemble possesses a repertoire of distinct output patterns. The decision on the choice of an appropriate (adaptive) pattern is determined by the balance of NTs in the ensemble extraneuronal milieu (or ECS).

6. NTs of input neurons participate in producing the ECS, thus making the ensemble activity dependent on the context.

## 4 Model formulation

### 4.1 Main definitions

Here we provide main definitions of the proposed model. A multi-transmitter neuronal system is a triple $\mathbf{S} = \langle \mathbf{N}, \mathbf{X}, \mathbf{C} \rangle$, where $\mathbf{N}=\{N_1, \ldots, N_n\}$ is a set of formal neurons, $\mathbf{X}$ – extracellular space (ECS) and $\mathbf{C}=\{c_1, \ldots, c_m\}$ is a set of neurotransmitters.

*Neural inputs.* Each neuron has several receptor slots, characterized by a transmitter $c_j$ and a weight $w_{ij} \in R$. A slot aggregates all receptors to a given transmitter and the weight represents the aggregated influence of this transmitter to the neuron. If $w_{ij}=0$ then transmitter $j$ has no effect on the neuron $i$. If $w_{ij}>0$ then transmitter $j$ excites neuron $i$ and $w_{ij}<0$ denotes that transmitter $j$ inhibits neuron $i$. Neuronal inputs are represented as a matrix $W = (w_{ij})_{n \times m}$.

*Neural outputs.* As we describe later the model functioning is divided into discrete times $t=1, 2,\ldots$ . Neuronal activity at time $t$ is denoted as $y_i(t) \in \{0, 1\}$; $y_i(t) = 1$ if neuron $i$ is active and firing spikes at time $t$. After an activation, a neuron releases some amount of one or several neurotransmitters. The neuronal outputs are represented as a matrix $D = (d_{ij})_{n \times m}$ where $d_{ij} \geq 0$ is the amount of transmitter $j$ released by neuron $i$.

*Extracellular space.* Neurons in the model communicate over the common extracellular space which contains the transmitters that have been released at time $t$. A state of ECS is represented as a vector $X(t) = (x_1(t), \ldots, x_m(t))$, where $x_j(t)$ denotes the amount of neurotransmitter $j$ present in the ECS at time $t$.

### 4.2 Excitation and inhibition

The neurons share the common ECS so every neuron is influenced by all transmitters to which it possesses receptors. For a compact description of the effects of NT on a neuron we define two technical concepts: excitation and inhibition functions. Each neuron has two thresholds: excitation threshold $P_{1i}$ and inhibition threshold $P_{0i}$, $P_{0i}<0<P_{1i}$. The excitation function $z_{1i}(t)$ indicates that neuron $i$ is excited at time $t$ and it is defined as follows:

$$z_{1i}(t) = \begin{cases} 1, \text{if } \sum_{j=1}^{m} w_{ij} x_j(t) \geq P_{1i} \\ 0 \text{ otherwise} \end{cases} \quad (1)$$

The inhibition function $z_{0i}(t)$, which indicates that a neuron is inhibited, is similar:

$$z_{0i}(t) = \begin{cases} 1, \text{if } \sum_{j=1}^{m} w_{ij} x_j(t) \leq P_{0i} \\ 0 \text{ otherwise} \end{cases} \quad (2)$$

Here $x_j(t)$, $l = 1, \ldots, m$ – the components of the ECS state at time $t$. One can see that it is possible that $z_{1i}(t)=z_{0i}(t)=0$, but it is impossible that $z_{1i}(t)=z_{0i}(t)=1$ since $P_{0i}<0$ and $P_{1i}>0$ for each neuron $i$.

## 4.3 Neuronal types

We consider three types of neurons which represent different membrane properties: oscillatory, tonic and passive follower. Each type is related to an individual firing behavior. Each neuron is represented as a finite automaton with two inputs and a set of internal states. So, the activity of a neuron at time $t$ is described by the following output function:

$$y_i(t) = F_{\theta(i)}(z_{1i}(t-1), z_{0i}(t), s_i(t-1)). \qquad (3)$$

Here $\theta(i)$ is a type of neuron $i$, $z_{1i}(t-1)$ is the excitation function's value at the previous time, $z_{0i}(t)$ is the inhibition function's value at time $t$ and $s_i(t-1)$ is the internal state at the previous time. One can see that the activity of neurons at time $t$ depends on the ECS state at the same time $t$. This contradictory definition is explained in the next section where we introduce the notion of neuronal competition. Each neuronal type is characterized by its own output function (3).

*Endogenous oscillator.* An endogenous oscillator or bursting neuron is characterized by endogenous bursts of spikes every $T_i$ times if not inhibited by other neurons at the moment of activation. This period of recharge may be different for different neurons. If an oscillatory neuron is inhibited it will be active after the inhibition disappears. If an oscillatory neuron is excited then it will become active at the next time immediately, not after the complete recharge period.

The internal structure and the output function of an oscillatory neuron are provided in the table 1.

| State | Inputs ($z_0, z_1$) | | |
|---|---|---|---|
| | ($z_0=0, z_1=0$) | ($z_0=1, z_1=0$) | ($z_0=0, z_1=1$) |
| $s_0$ | $s_1, y=0$ | $s_1, y=0$ | $s_0, y=1$ |
| … | … | … | … |
| $s_k$ | $s_{k+1}, y=0$ | $s_{k+1}, y=0$ | $s_0, y=1$ |
| … | … | … | … |
| $s_T$ | $s_0, y=1$ | $s_T, y=0$ | $s_0, y=1$ |

**Table 1:** State transitions and outputs of an oscillatory neuron

*Tonic neuron.* Neurons of this type are active as long as they are not inhibited, so tonic neurons have no internal state and their output function is as simple as that:

$$y_i(t) = \neg z_{0i}(t). \qquad (4)$$

*Follower neuron.* These neurons are active only after being excited by others so the output function takes the following form:

$$y_i(t) = \neg z_{0i}(t) z_{1i}(t-1). \qquad (5)$$

These definitions imply that tonic and follower neurons are degenerate cases of oscillatory neurons with zero and infinite recharge periods respectively. We distinguish these three types to clarify the distinct roles they occupy in a generator.

*Post-inhibitory rebound (PIR).* Besides the types described above we introduce post-inhibitory rebound gain coefficient that increase the output of a neuron that was inhibited at the previous time. Each neuron has an additional parameter – PIR gain coefficient $k_i^{PIR} \geq 1$. If the neuron was inhibited at time $t$ its output at the next time $t+1$ increases proportionally to PIR gain. The amount of neurotransmitter $j$ being released after an activation of neuron $i$ equals to $k_i^{PIR} z_{0i}(t-1) d_{ij}$.

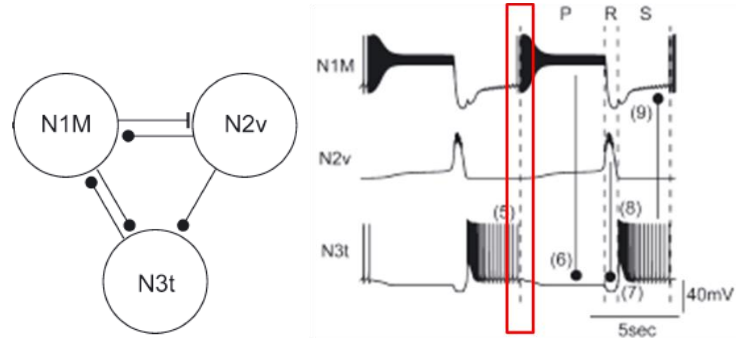

**Figure 1:** A competition between oscillatory N1M and tonic N3t neurons in the feeding CPG of the pond snail Lymnaea stagnalis **(Vavoulis, et al., 2007)**. These neurons inhibit each other as shown in the left picture. The dominant neuron defines the current phase of the feeding cycle

## 4.4 Neuronal competition and model dynamics

The main principle of the model's dynamics is a competition between neurons for the opportunity to be active during the next time step. In the model an inhibition is taking effect without a delay so we need a mechanism of conflict resolution to define what neurons can be active. In the paper the conflict resolution algorithm is introduced artificially. It doesn't pretend to mimic the competition in a biological CPG. But the purpose of the algorithm is to provide a reasonable conflict resolution so the whole model can generate rhythms that are similar to those observed in biological CPGs. Figure 1 shows an example from (Vavoulis, et al., 2007) where the competition plays a crucial role in the CPG.

The following algorithm is used to determine which neurons must be active during the next time $t$.

Algorithm 1. *Neuronal competition*
1. *Initialization of ECS.* Set the state of ECS to zero $x_j(t)=0$, $j=1,\ldots,m$
2. *Initialization of neurons.* Determine the potentially active neurons:
   $y_i(t)=F_i(z_{1i}(x(t-1)), 0, s_i(t-1))$.
3. *Update ECS.*

$$x_j(t) = \sum_{i=1}^{n} k^{PIR} z_{0i}(t-1) d_{ij} y_i(t)$$

4. *Conflict resolution.*
   a. Compute inhibition $z_{0i}(t)$ for each neuron
   b. If for each neuron $i \in N$ $z_{0i}(t)=0$ then go to step 5.
   c. Else find a neuron $k$ such that the value $\sum w_{ij} x_j(t) - P_{0i}$ is minimal among the neurons with $z_{0i}(t)=1$
   d. Set $y_k(t)=0$. This neuron cannot be active at time $t$.
   e. Go to step 3.
5. *Finish.* If there no more conflicts then the vectors $x(t)$, $y(t)$ is the states of the ECS and the neurons at time $t$.

The ECS and neurons' states are updated until the algorithm comes to a combination with no conflicts. During the steps 3 and 4 the number of active neurons can only decrease so the algorithm comes to a stable combination $x(t)$, $y(t)$ with at least one active neuron if the set of initially active neurons is not empty.

# 5  Examples

In this section, we provide two examples of CPGs described in terms of the proposed model. We consider a half-center oscillator (Marder & Bucher, 2001) and the three-phase feeding rhythms of the pond snail *Lymnaea stagnalis* (Vavoulis, et al., 2007).

## 5.1  Half-center oscillator

The example is based on papers (Marder & Bucher, 2001), (Marder, Goeritz, & Otopalik, 2015). The authors consider two basic mechanism underlying pattern production: pacemaker/follower and reciprocal inhibition. Figure 2 illustrates the examples. In the first case an endogenous oscillator drives other neurons into a rhythmic activity pattern. This mechanism can be trivially implemented by the model. The second case is less obvious. A rhythm emerges from mutual inhibition of two neurons which do not fire bursts when isolated but produce alternating phasic activity when coupled and inhibit each other. This type of network is called 'half-center oscillator' and is believed to be a common mechanism of pattern generation. Our model allows to formalize this mechanism and catch the essential role that is played by neuronal competition and post-inhibitory rebound.

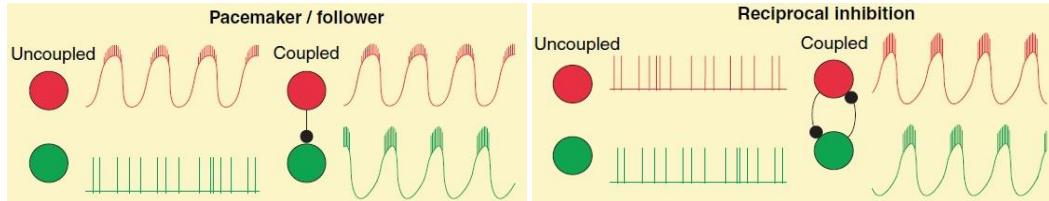

**Figure 2:** Basic mechanisms of rhythmic activity generation (**Marder & Bucher, Central pattern generators and the control of rhythmic movements, 2001**): (left) oscillations are driven by an endogenous oscillator – pacemaker; (right) oscillations emerge from reciprocal inhibition of two tonic neurons.

Half-center oscillator (HCO) consists of two neurons coupled by inhibitory links as shown in the right part of Figure 2. The inhibitory links are implemented via two different neurotransmitters, define them as transmitters *a* and *b*. Table 2 summarizes the parameters of the modeled half-oscillator. One of the neurons is assigned with more intensive output just to determine the winner of the first competition. PIR coefficients aimed to show that tonic firing is usually less intensive that phasic bursting activity. Excitation thresholds don't matter in that example and the inhibition thresholds are equal and set to -1.

| Neuron | Type | PIR gain | Thresholds | | Output $D$ | | Input $W$ | |
|---|---|---|---|---|---|---|---|---|
| | | | *Inhibition* $P_{0i}$ | *Excitation* $P_{1i}$ | $a$ | $b$ | $a$ | $b$ |
| $N_1$ | Tonic | 2 | -1 | - | 1.1 | 0 | 0 | -1 |
| $N_2$ | Tonic | 2 | -1 | - | 0 | 1 | -1 | 0 |

**Table 1:** Model of a half-center oscillator

Figure 3 illustrates the configuration of interactions in the model HCO and the produced rhythms. At the first time the neuron with maximal output won the competition, then the second neuron got its output increased due to rebound gain and won the competition at the second time. In the following times the rhythm is stable and the two neurons oscillate due to the competition and the effect of post-inhibitory rebound. This example shows the emergence of phasic oscillations in the absence of an endogenous oscillator.

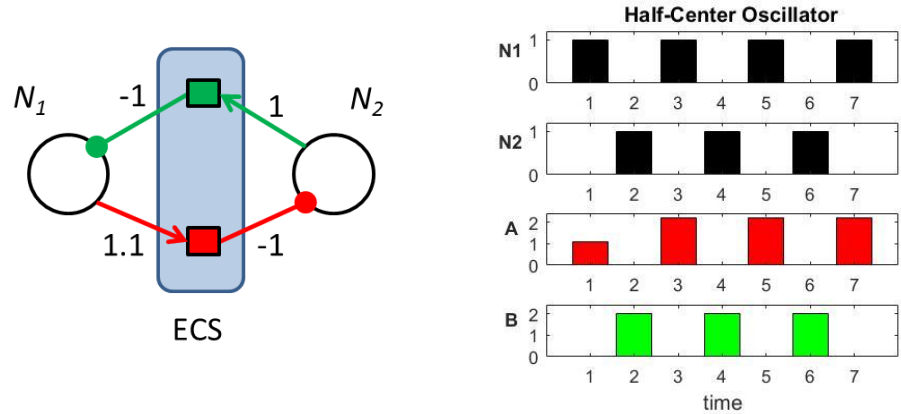

**Figure 3:** Model of a half-center oscillator: (left) structure of the neurons' interactions; (right) dynamics of the neurons' and ECS states

## 5.2 Snail feeding rhythm

The feeding generator of a pond snail *Lymnaea stagnalis* is a well-studied example of a central pattern generator. As shown in the Figure 1, the network consists of three interneurons responsible for separate phases of the feeding rhythm: protraction, rasp and swallow. In (Vavoulis, et al., 2007) a model based on Hodgkin & Huxley equations was proposed and it took the form of 38-dimensional system of ordinary differential equations which were solved numerically. Here we propose a simplified discrete model that emphasizes the logic of the interactions and neuronal roles in the CPG.

The network consists of three neurons $N_1$, $N_2$, $N_3$, each produces its own transmitter: *ach*, *glu* and *xxx* because the third transmitter in the feeding generator still has not been investigated experimentally and remains unknown. Let's consider what properties the neurons should possess to produce the forward feeding rhythm by the proposed model. $N_1$ neuron is an endogenous oscillator that drives the whole rhythm. $N_2$ is a follower that must be excited by $N_1$ to activate. $N_3$ is a tonic neuron with weak default output. The types are assigned according to the paper (Vavoulis, et al., 2007).

The network parameters are shown in the Table 3 and the produced rhythm is shown in Figure 4. The default output of $N_3$ is lower than that of $N_1$ so the first phase is won by $N_1$. Then $N_2$ wins the competition because of its high output. After being inhibited $N_3$ is able to win at the third time and drives the third phase of the rhythm. Then the gain effect of PIR disappears and $N_1$ wins the competition again. However, there are several combinations of model parameters that can produce the same rhythm. The question of how to choose the most efficient combination is left for further studies.

| Neuron | Type | PIR gain | Thresholds | | Output $D$ | | | Input, $W$ | | |
|---|---|---|---|---|---|---|---|---|---|---|
| | | | *Inhibition* $P_{0i}$ | *Excitation* $P_{1i}$ | ach | glu | xxx | ach | glu | xxx |
| $N_1$ | Oscillator | 1 | -1 | 1 | 1 | 0 | 0 | 0 | -1 | -1 |
| $N_2$ | Follower | 1 | -1 | - | 0 | 2 | 0 | 1 | 0 | 0 |
| $N_3$ | Tonic | 3 | -1 | - | 0 | 0 | 0.5 | -1 | -1 | 0 |

**Table 2:** Model of the feeding CPG

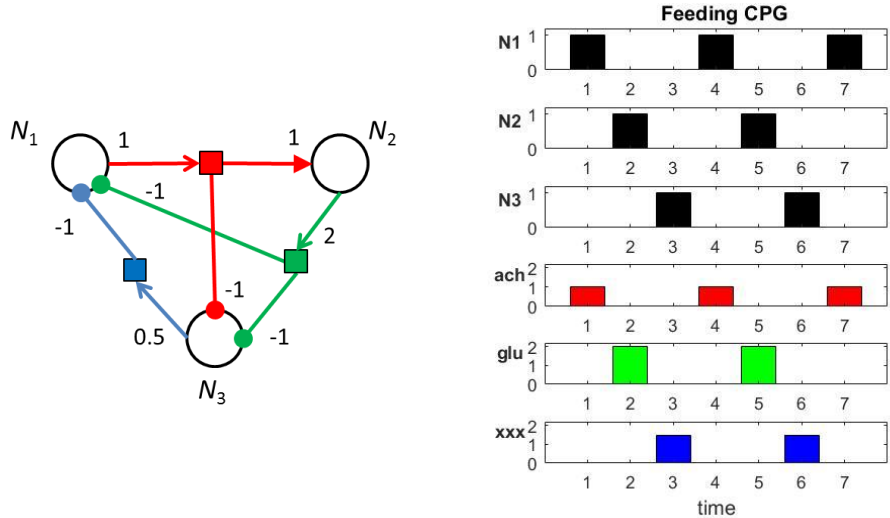

**Figure 4:** Feeding CPG model: (left) structure of neuronal interactions; (right) produced rhythm: neuronal activity (up) and concentrations of neurotransmitters (bottom)

# 6 Conclusion

We propose a formalized model of a multi-transmitter neural network where neurons interact via shared extracellular space without synaptic connections. Each neuron receives signals from the rest of the network by an individual set of receptors to a subset of the neurotransmitters which are released by other neurons. The model is intended to be a proof-of-concept example that some functional patterns of neural activity can in principle be implemented without synaptic wiring. In the model, we consider three various types of neurons differing in their electrical activity: tonic neurons, oscillators and followers. Oscillators and tonic neurons generate endogenous activity unless they are inhibited by other neurons. An algorithm of neuronal competition is introduced to resolve conflicts between those neurons that inhibits each other and are not allowed to be simultaneously active.

To illustrate the key features of the model we considered two examples of central pattern generators – neuronal ensembles that are able to generate rhythmic activity in the absence of an external input. The modeled generators are the well-known half-center oscillator and the feeding network of a pond snail *Lymnaea stagnalis*. The model is able to produce rhythms similar to those observed experimentally and in continuous modeling.

The proposed model, despite its simplicity, proved to be capable of simulation and explanation of the phenomena taking place in living neural ensembles.

# 7 References

Abbott, L. (1999). Lapique's introduction of the integrate-and-fire model neuron (1907). *Brain Research Bulletin, 50 (5/6)*, 303–304.

Balaban, P., Vorontsov, D., Dyakonova, V., Dyakonova, T., Zakharov, I., Korshunova, T., . . . Falikman, M. (2015). Central Pattern Generators. *Zhurn. Vyssh. Nerv. Deyat., 63*(5), 520–541.

Bargmann, C. (2012). Beyond the connectome: How neuromodulators shape neural circuits. *BioEssays, 34*(6), 458–465. Retrieved from Project "Open Connectome": http://www.openconnectomeproject.org/

Baronchelli, A., Ferrer-i-Cancho, R., Pastor-Satorras, R., Chater, N., & Christiansen, M. (2013). Networks in Cognitive Science. *Trends in Cognitive Sciences, 17*(7), 348–360.

Bloom, F. (1984). The functional significance of neurotransmitter diversity. *Am. J. Physiol., 246*, 184-94.

Brezina, V. (2010). Beyond the wiring diagram: signalling through complex neuromodulator networks. *Phil. Trans. R. Soc. Lond. B. Biol. Sci., 365*(1551), 2363-2374.

Bullmore, E., & Sporns, O. (2009). Complex brain networks: graph theoretical analysis of structural and functional systems. *Nature Reviews Neuroscience, 10*, 186-198.

Burks, A., & Wright, G. (1953). Theory of logical nets. *Proc. IRE., 41*(10), 1357–1365.

Dorogovtsev, S. (2010). *Lectures on Complex Networks.* Oxford: Oxford Univer. Press.

Dyakonova, V. (2012). Neurotransmitter mechanisms of context-dependent behavior. *Zh. Vyssh. Nerv. Deyat., 62*(6), 1–17.

FitzHugh, R. (1969). Mathematical models of excitation and propagation in nerve. In H. Schwan, *Biological Engineering* (pp. 1–85). N.Y.: McGraw–Hill Book Co.

Florey, E. (1967). Neurotransmitters and modulators in the animal kingdom. *Fed. Proc., 26*, 1164-1178.

Getting, P. (1989). Emerging principles governing the operation of neural networks. *Annu. Rev. Neurosci., 12*, 185–204.

Ghigliazza, R., & Holmes, P. (2004). A Minimal Model of a Central Pattern Generator and Motoneurons for Insect Locomotion. *SIAM J. Appl. Dyn. Syst., 3*(4), 671–700.

Ghigliazza, R., & Holmes, P. (2004). Minimal models of bursting neurons: The effects of multiple currents and timescales. *SIAM J. Appl. Dyn. Syst., 3*(4), 636–670.

Goodfellow, I., Bengio, Y., & Courville, A. (2016). *Deep Learning.* 787: MIT Press.

Graybiel, A. (1997). The basal ganglia and cognitive pattern generators. *Schizophr. Bull., 23*(3), 459–469.

Harris-Warrick, R. (2011). Neuromodulation and flexibility in Central Pattern Generator networks. *Curr. Opin. Neurobiol., 21*(5), 685–692.

Haykin, S. (2009). *Neural Networks and Learning Machines (3rd Edition).* Prentice-Hall.

Hodgkin, A. L., & Huxley, A. F. (1952). A quantitative description of membrane current and its applications to conduction and excitation in nerve. *J. Physiol. (Lond.), 116*, 500–544.

Hyafil, A., Fontolan, L., Kabdebon, C., Gutkin, B., & Giraud, A. (2015). Speech encoding by coupled cortical theta and gamma oscillations. *eLife*, 4:e06213.

Jarrell, T., Wang, Y., Bloniarz, A., Brittin, C., Xu, M., Thomson, J., . . . Emmons, S. (2012). The connectome of a decision-making neural network. *Science, 337*(6093), 437-444.

Kleene, S. (1956). Representation of Events in Nerve Nets and Finite Automata. In C. Shannon, & J. McCarthy, *Automata Studies.* Princeton: Princeton University Press.

Koch, C., & Segev, I. (1999). *Methods in neuronal modeling: from ions to networks (2nd ed.).* Cambridge, Massachusetts: MIT Press.

Kuznetsov, O. (2015). Complex networks and activity spreading. *Automation and Remote Control, 76*(12), 2091-2109.

LeCun, Y., Bengio, Y., & Hinton, G. (2015). Deep learning. *Nature, 521*(7553), 436–444.

Marder, E., & Bucher, D. (2001). Central pattern generators and the control of rhythmic movements. *Curr. Biol., 11*(23), R986–996.

Marder, E., Goeritz, M., & Otopalik, A. (2015). Robust circuit rhythms in small circuits arise from variable circuit components and mechanisms. *Curr. Opin. Neurobiol., 31*, 156-163.

McCulloch, W., & Pitts, W. (1943). A logical calculus of the ideas immanent in nervous activity. *Bull. Math. Biophys., 5*, 115-133.

Meyrand, P., Simmers, J., & Moulins, M. (1991). Construction of a pattern-generating circuit with neurons of different networks. *Nature, 351*, 60–63.

Minsky, M., & Papert, S. (1969). *Perceptrons, An Introduction to Computational Geometry.* Cambridge, Massachusetts: The MIT Press.

Moroz, L., & Kohn, A. (2016). Independent origins of neurons and synapses: insights from ctenophores. *Phil. Trans. R. Soc. Lond. B. Biol. Sci., 371*(1685).

Morris, C., & Lecar, H. (1981). Voltage Oscillations in the barnacle giant muscle fiber. *Biophys. J., 35 (1)*, 193–213.

Mulloney, B., & Smarandache, C. (2010). Fifty years of CPGs: two neuroethological papers that shaped the course of neuroscience. *Front. Behav. Neurosci., 4*(45), 1-8.

Nagumo, J., Arimoto, S., & Yoshizawa, S. (1962). An active pulse transmission line simulating nerve axon. *Proc. IRE, 50*, 2061–2070.

Rabinovich, M., Varona, P., Selverston, A., & Abarbanel, H. (2006). Dynamical principles in neuroscience. *Rev. Mod. Phys., 78*, 1213-1265.

Roberts, A., Conte, D. H.-H., Buhl, E., Borisyuk, R., & Soffe, S. (2014). Can Simple Rules Control Development of a Pioneer Vertebrate Neuronal Network Generating Behavior? *The Journal of Neuroscience, 34*(2), 608–621.

Roberts, P. (1998). Classification of Temporal Patterns in Dynamic Biological Networks. *Neural Computation, 10*(7), 1831-1846.

Rosenblatt, F. (1962). *Principles of Neurodynamic: Perceptrons and the Theory of Brain Mechanisms.* Washington, D.C.: Spartan books.

Sakharov, D. (1974). Evolutionary aspects of transmitter heterogeneity. *J. Neutal Transmission. Suppl. XI*, 41-59.

Sakharov, D. (1990). The multiplicity of neurotransmitters: the functional significance. *Zh. Evol. Biokhim. Fiziol., 26*(5), 733-741.

Sterratt, D., Graham, B., Gillies, A., & Willshaw, D. S. (2011). *Principles of computational modelling in neuroscience.* Cambridge University Press.

Tsukerman, V., & Kulakov, S. (2015). A temporal ratio model of the episodic memory organization in the ECI-networks. *Contemporary Engineering Sciences, 8*, 865–876.

Vavoulis, D., Straub, V., Kemenes, I., Kemenes, G., Feng, J., & P., B. (2007). Dynamic control of a central pattern generator circuit: a computational model of the snail feeding network. *European Journal of Neuroscience, 25*, 2805–2818.

Wang, R.-S., & Albert, R. (2013). Effects of community structure on the dynamics of random threshold networks. *Physical Review E., 87*, 012810.